\documentclass[pdflatex,sn-mathphys-num]{sn-jnl}

\usepackage{graphicx}%
\usepackage{multirow}%
\usepackage{amsmath,amssymb,amsfonts}%
\usepackage{bbm}
\usepackage{amsthm}%
\usepackage{mathrsfs}%
\usepackage[title]{appendix}%
\usepackage{xcolor}%
\usepackage{textcomp}%
\usepackage{manyfoot}%
\usepackage{booktabs}%
\usepackage{algorithm}%
\usepackage{algorithmicx}%
\usepackage{algpseudocode}%
\usepackage{listings}%
\usepackage{underscore}
\usepackage{graphicx}
\usepackage{booktabs}
\usepackage{siunitx}
\usepackage{pgfplotstable}
\usepackage{adjustbox}
\usepackage{subcaption}
\usepackage{textcomp}

\pgfplotsset{compat=1.18}
\usepackage{pgfplotstable}
\usepackage{booktabs}
\usepackage{textcomp} 
\usepackage{graphicx}
\usepackage{subcaption} 
\usepackage{booktabs}   
\usepackage{amsmath}    

\theoremstyle{thmstyleone}%
\theoremstyle{thmstyletwo}%

\theoremstyle{thmstylethree}%

\usepackage{lineno}

\begin{document}

\title[Representation geometry shapes task performance in vision-language modeling for CT enterography]{Representation geometry shapes task performance in vision-language modeling for CT enterography}


\author*[1]{\fnm{Cristian} \sur{Minoccheri}}\email{minoc@umich.edu}

\author[1]{\fnm{Emily} \sur{Wittrup}}\email{ewittrup@med.umich.edu}

\author[1,3,4]{\fnm{Kayvan} \sur{Najarian}}\email{kayvan@umich.edu}

\author[1,2]{\fnm{Ryan} \sur{Stidham}}

\affil[1]{\orgdiv{Gilbert S. Omenn Department of Computational Medicine and Bioinformatics},  \orgname{University of Michigan}, \orgaddress{\street{1109 Geddes Avenue}, \city{Ann Arbor}, \postcode{48104}, \state{MI}, \country{USA}}}

\affil[2]{\orgdiv{Department of Gastroenterology}, \orgname{University of Michigan}, \orgaddress{\street{1109 Geddes Avenue}, \city{Ann Arbor}, \postcode{48104}, \state{MI}, \country{USA}}}

\affil[3]{\orgdiv{Department of Emergency Medicine}, \orgname{University of Michigan}, \orgaddress{\street{1109 Geddes Avenue}, \city{Ann Arbor}, \postcode{48104}, \state{MI}, \country{USA}}}

\affil[4]{\orgdiv{Department of Electrical Engineering and Computer Science}, \orgname{University of Michigan}, \orgaddress{\street{1109 Geddes Avenue}, \city{Ann Arbor}, \postcode{48104}, \state{MI}, \country{USA}}}


\abstract{Computed tomography (CT) enterography is a primary imaging modality for assessing inflammatory bowel disease (IBD), yet the representational choices that best support automated analysis of this modality are unknown. We present the first study of vision-language transfer learning on abdominal CT enterography and identify two main findings. First, mean pooling of slice embeddings gives better categorical disease assessment (59.2\% three-class accuracy), whereas attention pooling gives better cross-modal retrieval (0.235 text-to-image MRR). This pattern holds across all LoRA configurations tested and suggests that the two aggregators emphasize different properties of the learned representation. Second, per-slice tissue contrast matters more than broader spatial coverage: multi-window RGB encoding, which maps complementary Hounsfield Unit windows to RGB channels, outperforms all strategies that increase spatial coverage through multiplanar sampling, and in this setting adding coronal and sagittal views reduces classification performance. For report generation, fine-tuning without retrieval context yields within-1 severity accuracy at the prevalence-matched chance level (70.4\% vs.\ 71\% random), suggesting little learned ordering beyond the class distribution. Retrieval-augmented generation (RAG) improves this across all configurations, scoring 7--14 percentage points above the chance baseline and improving ordinal MAE from 0.98 to 0.80--0.89. A three-teacher pseudolabel framework enables all comparisons without expert annotations. Together, these findings provide the first baselines for this underexplored modality and offer practical guidance for building vision-language systems for volumetric medical imaging.
}

\maketitle

\section{Introduction}

Inflammatory bowel disease (IBD), encompassing Crohn's disease and ulcerative colitis, affects approximately 3 million adults in the United States and represents a growing global health burden with rising incidence in newly industrialized countries \cite{dahlhamer2016ibd,ng2017worldwide}. The chronic, relapsing nature of IBD requires repeated imaging for disease monitoring, with computed tomography (CT) enterography serving as a primary modality for assessing transmural inflammation, detecting complications such as strictures and fistulae, and evaluating treatment response \cite{defined2019acr,defined2016esmrmr}. Radiological interpretation of these examinations demands integration of findings across multiple bowel segments and imaging planes, a cognitively demanding process subject to substantial inter-reader variability \cite{bhatnagar2022variability}. Automated analysis of CT enterography could address these challenges, but doing so requires learned representations that capture the right structure of the data -- and it is not obvious what that structure should be for different downstream tasks.

The intersection of computer vision and natural language processing (NLP) has yielded vision-language foundation models with remarkable capabilities for image understanding. Contrastive Language-Image Pre-training (CLIP) \cite{radford2021clip} demonstrated that contrastive learning on large-scale image-text pairs produces representations enabling zero-shot classification, cross-modal retrieval, and efficient transfer to downstream tasks. This paradigm has been adapted to the biomedical domain through models such as BiomedCLIP \cite{zhang2023biomedclip}, trained on 15 million scientific image-text pairs, and PubMedCLIP \cite{eslami2023pubmedclip}. These models leverage the abundant paired image-text data in scientific literature and clinical archives to learn generalizable representations without task-specific supervision.

The application of vision-language models to radiology has achieved notable success, though predominantly focused on chest X-ray interpretation. CheXzero \cite{tiu2022chexzero} demonstrated that CLIP-style contrastive learning on chest radiographs and reports enables zero-shot pathology detection competitive with supervised approaches. The MIMIC-CXR dataset \cite{defined2019mimiccxr}, containing over 370,000 chest radiographs with associated reports, has enabled extensive model development and benchmarking. These successes have established chest radiography as a proving ground for medical vision-language learning -- but they have also created an implicit assumption that methods and design choices developed for 2D chest images will transfer to other modalities.

This assumption deserves scrutiny for volumetric imaging. Three-dimensional CT presents fundamental representational challenges: the dimensionality mismatch with 2D pre-trained encoders, substantially larger data volumes, and complex spatial relationships that extend across the imaging volume. CT-CLIP \cite{hamamci2024ctclip} addressed this by training a dedicated 3D vision encoder on 104,000 chest CT volumes, while RadFM \cite{wu2023radfm} and M3D \cite{defined2024m3d} employed perceiver-based and large-scale multimodal approaches. These methods require substantial computational resources, motivating parameter-efficient adaptation strategies that leverage pre-trained 2D encoders. But beyond computational efficiency, a more fundamental question arises: for a given volumetric modality and clinical task, what representational choices actually matter?

Abdominal CT enterography is a natural testbed for this question. The greater anatomical complexity and wider range of pathological processes affecting the gastrointestinal tract create a demanding learning problem. Unlike chest CT, no large public benchmarks exist for CT enterography, and to our knowledge no prior work has applied vision-language contrastive learning to abdominal CT with paired radiology reports. This gap means that basic representational questions -- how to encode 3D volumes as 2D inputs, how to aggregate slice-level features, how to handle templated text -- must be answered from first principles.

This work presents a systematic study of representational choices for CT enterography vision-language learning, evaluated across three tasks: disease activity classification, cross-modal retrieval, and automated impression generation. We adapt BiomedCLIP \cite{zhang2023biomedclip} for volumetric imaging through a 2.5D slice-based approach, exploring a range of input encoding strategies, slice aggregation methods, and LoRA configurations \cite{hu2021lora}. A multi-positive contrastive loss formulation addresses the many-to-one mapping between CT scans and templated impression text. For report generation, we fine-tune MedGemma \cite{sellergren2024medgemma} and explore retrieval-augmented generation (RAG) using learned embeddings. Given the absence of expert annotations, we develop a three-teacher ensemble combining rule-based clinical NLP \cite{chapman2001negex} with two open-source LLMs \cite{labrak2024biomistral,yang2024qwen2} to generate consensus pseudolabels -- a practical framework that enables rigorous methodological comparison from routine clinical archives without expert labeling.

Our main finding is that the embedding geometry that best supports categorical disease assessment (mean pooling, which produces globally consistent features) differs from the geometry that best supports cross-modal retrieval (attention pooling, which preserves slice-specific information). This pattern holds across all LoRA configurations tested and also affects which embeddings are most useful for RAG-based report generation. The fact that simpler design choices (mean pooling, per-slice multi-window encoding) outperform more complex alternatives (transformer aggregators, multiplanar sampling) further suggests that for categorical assessment in limited-data settings, representation richness matters more than architectural complexity.

The main contributions of this paper are:

1) A systematic analysis of representation geometry for CT enterography vision-language learning, showing a consistent classification--retrieval trade-off across aggregation strategies and LoRA configurations that informs application-specific model selection;

2) A practical 2.5D BiomedCLIP adaptation establishing that multi-window RGB encoding and mean pooling are strong defaults for classification (59.2\% three-class accuracy), while attention pooling is preferred for retrieval (0.235 MRR) -- the first baselines for this underexplored modality; and

3) Evidence that fine-tuning without retrieval context does not preserve severity ordering beyond the prevalence-matched chance level, whereas RAG scores 7--14 percentage points above that baseline across all embedding configurations tested in this limited-data setting.

Experiments on 1,074 CT enterography examinations show that the classification--retrieval trade-off appears across all configurations tested, that per-slice tissue contrast via multi-window RGB encoding matters more than broader spatial coverage, and that retrieval-augmented generation reduces the severity confusion seen in the fine-tuned baseline. These results provide the first baselines for this modality and practical guidance for practitioners building vision-language systems on volumetric medical imaging. The code used to generate these results is available at https://github.com/Minoch/RadIBD.

\section{Dataset}

The study utilized CT enterography examinations performed at the University of Michigan. Initial data extraction identified 1,422 CT enterography studies with paired radiology reports containing both findings and impression sections.

DICOM series were ordered and used to build CT volumes. Raw pixel values were converted to HU using DICOM rescale parameters and clipped to $[-1000, 1000]$ HU. Volumes with fewer than 30 slices were excluded to ensure adequate anatomical coverage, and a series selection policy retained only the largest series per examination.
Isotropic resampling to $1.0$ mm$^3$ spacing was performed using trilinear interpolation via SimpleITK \cite{lowekamp2013simpleitk}. The HU clipping range of $[-1000, 1000]$ was selected to encompass relevant soft tissue and pathological findings while excluding extreme values \cite{roth2014lymph}.

Disease activity labels were generated through a three-teacher ensemble approach combining rule-based clinical NLP with large language models, as detailed in the Methods section. Of the initial 1,422 studies with paired reports, 348 were excluded due to abstention by the consensus labeling system (all three teachers disagreed) or insufficient report text for reliable classification, yielding a final cohort of 1,074 CT enterography examinations. This filtering ensured that all retained studies had at least majority agreement (two of three teachers) on disease activity classification, providing more reliable training signal despite the absence of expert annotations.

The final dataset was split into training ($n=839$, 78.1\%), validation ($n=110$, 10.2\%), and test ($n=125$, 11.6\%) sets using patient-level stratification to prevent data leakage from repeated examinations of the same individual. The test set label distribution reflected the clinical population, comprising 39 normal (31.2\%), 28 possibly abnormal (22.4\%), and 58 abnormal (46.4\%) cases based on consensus pseudolabels. Among test set labels, 35 (28.0\%) achieved unanimous agreement across all three teachers (high confidence), while 90 (72.0\%) achieved majority agreement (medium confidence). This confidence stratification enables downstream analysis of model performance as a function of label certainty.

The dataset presented several characteristics that directly shape the experimental design. First, impression text exhibited substantial redundancy: many normal studies share identical or near-identical templated language (e.g., ``no evidence of active inflammatory bowel disease''). This many-to-one mapping between CT volumes and impression text motivates the multi-positive contrastive loss formulation and also creates an inherent asymmetry in the retrieval task -- multiple scans share the same text, making text-to-image retrieval easier than image-to-text retrieval. Second, the three-class label distribution was imbalanced toward abnormal cases (46.4\% in the test set), reflecting the referral population undergoing CT enterography for suspected or known IBD; random baselines must account for this imbalance. Third, reports varied considerably in length and detail, ranging from single-sentence impressions for straightforward normal studies to multi-paragraph descriptions for complex cases with multiple findings.

\section{Methods}

The experimental design is structured around a single question: which representational choices determine task performance for CT enterography vision-language learning, and do those choices differ across tasks? We evaluate three tasks -- disease activity classification, cross-modal retrieval, and impression generation -- under the same supervision framework, enabling direct comparison of how the learned representations affect each.

\subsection*{Pseudolabel generation}

Given the absence of expert annotations, we employed a three-teacher ensemble to generate consensus pseudolabels. Reports were classified into three categories: normal (no clinically meaningful abnormality), abnormal (definite active disease including inflammation, infection, obstruction, perforation, abscess, or fistula), and possibly abnormal (uncertain or hedged findings, limited study quality, or historical findings only). This taxonomy is impression-level and report-driven, designed to capture the dominant severity signal in routine radiologist language rather than an adjudicated clinical activity score.

The first teacher implemented a NegEx/ConText-style rule-based classifier \cite{chapman2001negex,harkema2009context} applied to both impression and findings sections, identifying negation patterns (``no,'' ``without,'' ``absence of''), uncertainty markers (``possible,'' ``may represent,'' ``cannot exclude''), historical context indicators (``history of,'' ``prior,'' ``chronic''), and present/acute indicators (``active,'' ``acute,'' ``flare''). Clinical concepts covered IBD-specific inflammation terms (colitis, enteritis, ileitis), objective imaging findings (wall thickening, mural enhancement, submucosal edema), and complications (abscess, fistula, stricture, perforation).

Two open-source large language models served as additional teachers: BioMistral-7B \cite{labrak2024biomistral}, a biomedically-adapted Mistral model, and Qwen2.5-7B-Instruct \cite{yang2024qwen2}, a general-purpose instruction-tuned model. Both were prompted with identical few-shot instructions and classification rules, with generation performed deterministically. Final labels were assigned by majority voting: high confidence when all three teachers agreed, medium confidence for two-of-three agreement, and abstention when all three disagreed (these cases were excluded from the dataset).

\subsection*{CT volume encoding and aggregation}

We adapted BiomedCLIP \cite{zhang2023biomedclip}, pre-trained on 15 million biomedical image-text pairs, which uses a Vision Transformer (ViT-B/16) \cite{dosovitskiy2020vit} for image encoding and PubMedBERT \cite{gu2021pubmedbert} for text encoding, producing 512-dimensional embeddings in a shared latent space.

To encode 3D CT volumes with a 2D vision encoder, we used a 2.5D slice-based approach. Our primary configuration extracted 16 axial, 6 coronal, and 6 sagittal slices using linearly-spaced sampling from 20\% to 80\% of the volume extent. Three complementary HU windows were mapped to RGB channels: $[-150, 250]$ HU for soft tissue (red), $[-1000, 1000]$ HU for full dynamic range (green), and $[0, 500]$ HU for enhanced structures (blue). This multi-window strategy encodes diverse tissue contrasts within a single image compatible with pre-trained vision encoders.

To isolate the contribution of each design choice, we ablated three dimensions: (1) plane sampling -- axial-only versus multiplanar (axial, coronal, sagittal); (2) slice selection -- linearly-spaced versus stratified sampling; and (3) RGB encoding -- grayscale (single HU window replicated), adjacent-slice RGB (three consecutive slices mapped to R, G, B), or multi-window RGB (three HU windows on a single slice). These ablations directly address whether performance derives from spatial coverage, sampling strategy, or per-slice contrast richness.

Individual slice embeddings $\mathbf{e}_1, \ldots, \mathbf{e}_S \in \mathbb{R}^{512}$ were aggregated into a single volume embedding via one of three strategies: mean pooling, attention pooling with a learnable query vector ($\alpha_i = \text{softmax}(\mathbf{q}^\top \mathbf{e}_i / \sqrt{d})$), or a single-layer transformer encoder with learnable CLS token and positional embeddings. The choice of aggregation strategy is the primary variable driving the classification--retrieval trade-off examined in this paper. A residual projector with layer normalization, GELU activation, and dropout was applied to the aggregated embedding before contrastive loss computation.

\subsection*{Parameter-efficient fine-tuning and contrastive loss}

Low-Rank Adaptation (LoRA) \cite{hu2021lora} was used to adapt both vision and text encoders, decomposing weight updates as $W' = W + BA$ where $B \in \mathbb{R}^{d \times r}$, $A \in \mathbb{R}^{r \times k}$, and $r \ll \min(d, k)$. We varied LoRA rank and the number of adapted blocks across configurations (denoted vXbY\_tZbW for vision rank X over Y blocks, text rank Z over W blocks) to assess whether capacity interacts with aggregation strategy. We tested additional configurations including symmetric text block counts, but report only the best-performing subset as other configurations yielded lower performance.

Standard contrastive learning treats each sample as having a unique positive match, creating false negatives when multiple scans share identical impression text -- a pervasive feature of clinical radiology archives where normal studies share templated language. We therefore implemented multi-positive contrastive loss treating all samples with the same normalized impression as valid positives:
\begin{equation}
\mathcal{L} = -\frac{1}{2N}\sum_{i=1}^{N}\left[\log\frac{\sum_{j \in P_i} \exp(s_{ij}/\tau)}{\sum_{k=1}^{N}\exp(s_{ik}/\tau)} + \log\frac{\sum_{j \in P_i} \exp(s_{ji}/\tau)}{\sum_{k=1}^{N}\exp(s_{jk}/\tau)}\right]
\end{equation}
where $s_{ij} = \mathbf{v}_i^\top \mathbf{t}_j$ is the cosine similarity between volume and text embeddings, $P_i$ is the set of indices with matching normalized impression text, and $\tau$ is a learned temperature parameter.

\subsection*{Training and evaluation}

Models were trained for 10 epochs using AdamW optimizer with learning rate $5 \times 10^{-5}$, weight decay $10^{-2}$, and batch size of 8. Automatic mixed precision training with gradient clipping (norm $\leq 1.0$) was employed, and early stopping with patience of 3 epochs monitored validation loss. We ran configurations multiple times and observed minimal run-to-run variability; for clarity, we report representative best runs per configuration.

Cross-modal retrieval performance was evaluated on the test set for both image-to-text (ranking impression embeddings given a CT volume embedding) and text-to-image (ranking CT embeddings given an impression embedding) tasks. Recall@$K$ measures the fraction of queries where the correct match appears in the top $K$ retrieved results. Mean Reciprocal Rank (MRR) averages the inverse rank of the first correct result across queries, rewarding models that rank correct matches higher (MRR $= 1$ if always ranked first, approaching $0$ for random ranking). To handle duplicate impressions, retrieval was considered successful if any sample with matching normalized impression text appeared in the top-$K$ results.

For classification evaluation, a logistic regression classifier was trained on CT volume embeddings from the training set to predict consensus disease activity labels, using L2 regularization ($C=1.0$), balanced class weights, and multinomial formulation. Evaluation metrics included per-class precision, recall, and F1-score, overall accuracy, macro-averaged F1, and confusion matrix analysis across the three-class taxonomy.

\subsection*{Impression generation and RAG}

MedGemma-4B \cite{sellergren2024medgemma}, a medical vision-language model based on the Gemma architecture, was adapted for CT report generation. The model accepts interleaved image and text inputs through an AutoModelForImageTextToText architecture with unified tokenization.

Three-dimensional CT volumes were converted to 2D montage images by extracting 16 slices from axial, 10 coronal, and 10 sagittal planes using linear sampling from 20\% to 80\% of each axis. HU values were windowed to $[-160, 240]$ and normalized to $[0, 255]$, with three consecutive slices $(i-1, i, i+1)$ mapped to RGB channels to encode spatial context. Slices were resized to $256 \times 256$ pixels, arranged in a grid of 3 columns, and plane-specific montages were concatenated vertically, with final images resized to maximum 1536 pixels on the longest side. Train-time augmentation included slice index jitter ($\pm 3$ slices) and HU window jitter ($\pm 25$ HU).

Parameter-efficient fine-tuning was performed using LoRA \cite{hu2021lora} applied to both vision encoder and language model decoder, with rank $r=16$, scaling factor $\alpha=32$, and dropout 0.05. The model was fine-tuned to generate impression text given the montage image and a structured prompt formatted as a chat template with system message (``You are an expert abdominal radiologist''), user message containing the image and instructions, and assistant message containing the ground-truth impression. Loss was computed only on assistant tokens, with prompt tokens masked.

Training used AdamW optimizer with learning rate $7 \times 10^{-5}$, weight decay 0.01, linear warmup over 3\% of steps, batch size 1 with gradient accumulation over 16 steps, and gradient checkpointing. Early stopping with patience 3 monitored validation loss over up to 10 epochs. At inference, impressions were generated with maximum 240 and minimum 48 new tokens, sampling with temperature 0.6, top-$p$ 0.9, repetition penalty 1.08, and no-repeat 3-gram constraint. Best-of-4 candidate selection with up to 3 retries was employed for quality filtering, rejecting outputs with fewer than 30 characters or fewer than 1 sentence.

For comparison with the fine-tuned MedGemma model, we implemented retrieval-augmented generation (RAG) using CT-CLIP embeddings. Given a test volume, the volume embedding was compared against all training set embeddings via cosine similarity, and the top-5 most similar training impressions were retrieved. Optional Maximal Marginal Relevance (MMR) diversification with pool size 50 and $\lambda=0.7$ was applied to balance relevance and diversity. Retrieved impressions were provided as context to MedGemma-4B with explicit instructions not to copy examples and to generate 3--5 sentences focusing on IBD activity assessment.

Generated impressions were evaluated against ground-truth radiologist impressions using multiple complementary metrics, chosen because standard n-gram metrics and clinical severity metrics can diverge substantially -- a generated report that uses different phrasing but conveys the same disease activity may score poorly on BLEU while being clinically equivalent. ROUGE-L F1 measured longest common subsequence-based lexical overlap. BLEU computed n-gram precision with brevity penalty using SacreBLEU at sentence level. METEOR calculated harmonic mean of unigram precision and recall with stemming and synonymy matching using NLTK -- the only standard metric that rewards synonymous phrasing. BERTScore measured contextual embedding similarity using RoBERTa-large \cite{liu2019roberta}. To assess clinical severity fidelity, we applied the rules-based classifier to generated impressions and compared predicted labels against ground-truth consensus labels, reporting exact accuracy, ordinal MAE, and within-1 accuracy. For within-1 accuracy, we note that with three ordered classes the metric is only violated by maximum-distance errors (predicting normal when the true label is abnormal, or vice versa). Given the test set distribution (31\% normal, 22\% possibly abnormal, 46\% abnormal), a prevalence-matched random classifier achieves approximately 71\% within-1 by chance, and a uniform random classifier achieves 74\%; these provide the appropriate reference points for interpreting results. We also note that a degenerate always-predict-middle-class strategy achieves 100\% within-1, so this metric should be interpreted alongside ordinal MAE rather than in isolation.

\section{Results}

\subsection*{Input encoding: tissue contrast matters more than spatial coverage}

We first compared input encoding strategies to determine what information is most useful for CT volume representation (Table~\ref{tab:input_ablation}). Multi-window RGB encoding, which maps three complementary HU windows (soft tissue, wide-range, and high-attenuation) to RGB channels, achieved the best classification performance (56.8\% accuracy, 0.555 macro-F1) among all configurations tested under the baseline mean pooling and v4b6 LoRA capacity. Importantly, adding multiplanar sampling -- which provides coronal and sagittal views in addition to axial -- did not improve and in fact hurt performance at this capacity level, with the multiaxial configuration scoring 7.8 percentage points below axial-only with multi-window encoding. Stratified sampling also failed to help relative to linear spacing. These results consistently indicate that enriching the tissue contrast information in each slice is more valuable than increasing spatial coverage across the volume at this capacity regime. All subsequent experiments fix multi-window RGB encoding and use multiplanar sampling, which is the configuration used in Tables~\ref{tab:classification} and~\ref{tab:retrieval}; we note that at higher LoRA capacities the combination does not recover the performance gap seen in Table~\ref{tab:input_ablation}, consistent with the view that spatial coverage is not the binding constraint in this setting.

\begin{table}[htbp]
\centering
\caption{Input-encoding and sampling ablations using mean pooling aggregation and v4b6\_t4b6 LoRA configuration. Adjacent RGB maps consecutive slices to RGB; Multiwindow RGB maps HU windows to RGB; Axial uses single plane; Multiaxial uses axial, coronal, and sagittal planes. Random baseline: 33\% accuracy, 0.33 macro-F1.}
\label{tab:input_ablation}
\begin{tabular}{lcc}
\toprule
\textbf{Configuration} & \textbf{Accuracy} & \textbf{Macro-F1} \\
\midrule
Multiwindow RGB & \textbf{0.568} & \textbf{0.555} \\
Axial baseline & 0.538 & 0.523 \\
Adjacent RGB & 0.514 & 0.505 \\
Axial + stratified sampling & 0.506 & 0.489 \\
Multiaxial & 0.490 & 0.470 \\
Multiaxial + stratified sampling & 0.490 & 0.472 \\
Multiaxial + multiwindow & 0.474 & 0.468 \\
\bottomrule
\end{tabular}
\end{table}

\subsection*{The classification--retrieval trade-off}

Having fixed the input encoding, we evaluated aggregation strategies and LoRA configurations across both classification (Table~\ref{tab:classification}) and retrieval (Table~\ref{tab:retrieval}) tasks simultaneously, as this joint view reveals the main pattern in the paper.

For classification, mean pooling consistently outperformed attention-based aggregation at matched LoRA capacity (59.2\% vs.\ 55.2\% for the largest configuration, v8b12\_t8b6). The lightweight transformer aggregator performed worse still (51.2\%), and a deeper two-layer transformer yielded even lower performance (not shown). Increasing LoRA capacity helps within each aggregation family but does not close the gap between families. Per-class F1 scores show that abnormal cases were most reliably detected (0.655), while the intermediate ``possibly abnormal'' category proved most challenging (0.545), consistent with its inherent ambiguity in radiologist language. The best-performing model trained only 0.74M parameters while keeping the base BiomedCLIP encoders frozen, demonstrating that effective domain transfer does not require large-scale fine-tuning.

\begin{table}[htbp]
\centering
\caption{Three-class disease activity classification on the test set ($n=125$). All models use multiwindow RGB encoding with multiplanar sampling. Aggregator: Mean = mean pooling, Attn = attention pooling, Lite = lightweight transformer. LoRA: vXbY\_tZbW = vision rank X over Y blocks, text rank Z over W blocks. Random baseline: 33\% accuracy, 0.33 macro-F1.}
\label{tab:classification}
\begin{tabular}{lccccc}
\toprule
\textbf{Configuration} & \textbf{Accuracy} & \textbf{Macro-F1} & \textbf{F1 (Normal)} & \textbf{F1 (Poss.)} & \textbf{F1 (Abn.)} \\
\midrule
Mean, v8b12\_t8b6 & \textbf{0.592} & \textbf{0.580} & \textbf{0.541} & \textbf{0.545} & \textbf{0.655} \\
Mean, v4b6\_t4b6 & 0.568 & 0.555 & 0.514 & 0.515 & 0.636 \\
Mean, v8b6\_t4b6 & 0.536 & 0.523 & 0.480 & 0.478 & 0.611 \\
\midrule
Attn, v8b12\_t8b6 & 0.552 & 0.541 & 0.521 & 0.486 & 0.617 \\
Attn, v4b6\_t4b6 & 0.560 & 0.550 & 0.526 & 0.508 & 0.617 \\
Attn, v8b6\_t4b6 & 0.536 & 0.519 & 0.480 & 0.448 & 0.630 \\
\midrule
Lite, v4b6\_t4b6 & 0.512 & 0.495 & 0.475 & 0.413 & 0.598 \\
\bottomrule
\end{tabular}
\end{table}

The retrieval results reveal the other side of the trade-off. Attention pooling with v4b6\_t4b6 achieved the best text-to-image retrieval (MRR 0.235), while the best classification model (Mean, v8b12\_t8b6) achieved only MRR 0.166 -- a relative deficit of 29\% despite having greater LoRA capacity. Conversely, the best retrieval model achieved only 56.0\% classification accuracy. This inverse relationship is unlikely to be a random fluctuation: it holds consistently across \textit{all} LoRA configurations in both directions -- every mean pooling variant matches or outperforms its attention pooling counterpart at matched capacity for classification, and every attention pooling variant outperforms its mean pooling counterpart for retrieval. This cross-configuration consistency across six independent comparisons suggests that the pattern reflects a property of the aggregation geometry rather than configuration-specific noise, and provides some reassurance despite the modest test set size. Mean pooling's globally averaged representation provides stable, category-consistent features well-suited for separating disease severity classes, while attention pooling's slice-weighted representation preserves local detail useful for matching specific volume-report pairs.

\begin{table}[htbp]
\centering
\caption{Cross-modal retrieval performance on the test set ($n=125$). All models use multiwindow RGB encoding with multiplanar sampling. Aggregator: Mean = mean pooling, Attn = attention pooling, Lite = lightweight transformer. LoRA: vXbY\_tZbW = vision rank X over Y blocks, text rank Z over W blocks. I2T = image-to-text, T2I = text-to-image. Random baseline: R@1=0.008, R@5=0.04, R@10=0.08.}
\label{tab:retrieval}
\begin{tabular}{lcccc|cccc}
\toprule
& \multicolumn{4}{c|}{\textbf{Image-to-Text (I2T)}} & \multicolumn{4}{c}{\textbf{Text-to-Image (T2I)}} \\
\textbf{Configuration} & R@1 & R@5 & R@10 & MRR & R@1 & R@5 & R@10 & MRR \\
\midrule
Mean, v8b6\_t4b6 & \textbf{0.088} & 0.136 & 0.232 & 0.140 & \textbf{0.160} & 0.264 & \textbf{0.400} & 0.231 \\
Mean, v8b12\_t8b6 & 0.064 & \textbf{0.176} & \textbf{0.256} & 0.132 & 0.064 & 0.296 & 0.392 & 0.166 \\
Mean, v4b6\_t4b6 & 0.064 & 0.168 & 0.240 & 0.135 & 0.152 & \textbf{0.304} & 0.360 & 0.228 \\
\midrule
Attn, v4b6\_t4b6 & \textbf{0.088} & 0.160 & 0.248 & \textbf{0.146} & \textbf{0.160} & \textbf{0.304} & \textbf{0.400} & \textbf{0.235} \\
Attn, v8b6\_t4b6 & \textbf{0.088} & 0.168 & 0.216 & 0.142 & 0.064 & 0.280 & 0.392 & 0.169 \\
Attn, v8b12\_t8b6 & 0.064 & 0.160 & 0.232 & 0.128 & 0.152 & 0.272 & 0.368 & 0.228 \\
\midrule
Lite, v4b6\_t4b6 & 0.040 & 0.088 & 0.208 & 0.092 & 0.024 & 0.168 & 0.280 & 0.107 \\
\bottomrule
\end{tabular}
\end{table}

Text-to-image retrieval consistently outperformed image-to-text retrieval (MRR 0.235 vs 0.146 for Attn v4b6\_t4b6), reflecting the many-to-one mapping where multiple CT scans share similar impression text and making it easier to find a matching scan given text than vice versa. The lightweight transformer aggregator underperformed both simpler alternatives across both retrieval directions, reinforcing that complexity does not help in this limited-data setting.

\subsection*{Report generation: what does RAG actually provide?}

Having established the trade-off, we can now ask what it implies for report generation. We fine-tuned MedGemma \cite{sellergren2024medgemma} on CT montage images and compared it against RAG variants using CT-CLIP embeddings as retrieval indices (Table~\ref{tab:generation}). For RAG, lighter LoRA configurations suffice since generation quality depends primarily on the retrieved context rather than embedding precision.

Standard lexical metrics show a pattern that is largely interpretable rather than contradictory. RAG models achieve comparable ROUGE-L (0.114--0.124 vs.\ 0.126) and substantially higher METEOR (0.21 vs.\ 0.17) but much lower BLEU (1.3--1.5 vs.\ 6.3) than the fine-tuned baseline. The METEOR advantage reflects RAG's use of synonymous and paraphrastic phrasing common in radiologist language -- clinically equivalent expressions that METEOR's stemming and synonymy matching rewards but BLEU's strict n-gram matching penalizes. Conversely, the fine-tuned baseline's higher BLEU reflects partial memorization of templated phrases from training, the same redundancy that motivated the multi-positive contrastive loss -- this is not a quality advantage. BERTScore remains stable across all models ($\sim$0.83), confirming that contextual semantic similarity is preserved regardless of surface lexical differences. The near-identical performance across RAG configurations is itself a positive finding: retrieval quality saturates quickly, meaning practitioners can use lighter embedding models without sacrificing generation quality.

\begin{table}[htbp]
\centering
\caption{Report generation evaluation on the test set ($n=125$). RAG models use lighter LoRA configurations than classification models, as generation quality depends primarily on retrieval context. Aggregator: Mean = mean pooling, Attn = attention pooling, Lite = lightweight transformer. LoRA: vXbY\_tZbW = vision rank X over Y blocks, text rank Z over W blocks. BLEU is sentence-level SacreBLEU (0--100 scale).}
\label{tab:generation}
\begin{tabular}{lcccc}
\toprule
\textbf{Model} & \textbf{ROUGE-L} & \textbf{METEOR} & \textbf{BLEU} & \textbf{BERTScore} \\
\midrule
MedGemma (SFT) & \textbf{0.126} & 0.173 & \textbf{6.28} & 0.826 \\
\midrule
\multicolumn{5}{l}{\textit{RAG with CT-CLIP retrieval ($k=5$):}} \\
Mean, v4b3\_t4b3 & 0.121 & 0.216 & 1.46 & 0.828 \\
Mean, v4b6\_t4b6 & 0.122 & 0.214 & 1.45 & \textbf{0.830} \\
Attn, v4b3\_t4b3 & 0.119 & 0.215 & 1.43 & 0.828 \\
Attn, v4b6\_t4b6 & \textbf{0.124} & \textbf{0.218} & 1.42 & 0.828 \\
Lite, v4b6\_t4b6 & 0.115 & 0.209 & 1.36 & 0.828 \\
Lite, v4b3\_t4b3 & 0.114 & 0.210 & 1.29 & 0.827 \\
\bottomrule
\end{tabular}
\end{table}

To assess whether generated impressions preserve the clinical severity signal, we applied the rules-based classifier to generated text and compared predicted labels against ground-truth consensus labels (Table~\ref{tab:label_consistency}). We report ordinal MAE (treating normal=0, possibly abnormal=1, abnormal=2) and within-1 accuracy alongside standard classification metrics.

The clearest result in this section is the contrast between the fine-tuned baseline and RAG. With three ordered classes, within-1 accuracy is only violated by the maximum possible error -- predicting ``normal'' when the true label is ``abnormal'' or vice versa. The appropriate random baselines for our test set distribution (31\% normal, 22\% possibly abnormal, 46\% abnormal) are 71\% for prevalence-matched random guessing and 74\% for uniform random guessing. The fine-tuned MedGemma baseline scores 70.4\% -- essentially at the prevalence-matched chance level and below uniform random -- indicating that fine-tuning without retrieval context fails to learn any severity-ordering signal beyond what the class distribution alone would predict. RAG models score 78--85\%, consistently above both random baselines, with ordinal MAE improving from 0.98 to 0.80--0.89. This 7--14 percentage point improvement over prevalence-matched chance is consistent across all six RAG configurations and indicates better preservation of severity ordering rather than a marginal fluctuation. In this setting, the main contribution of RAG is to reduce the severity-ordering errors seen in the fine-tuned baseline, not merely to nudge text-overlap metrics upward.

\begin{table}[htbp]
\centering
\caption{Label consistency evaluation: accuracy of disease activity labels extracted from generated impressions compared to ground-truth consensus labels. Models match those in Table~\ref{tab:generation}. Ordinal metrics treat labels as ordered (normal $<$ possibly abnormal $<$ abnormal). Random baselines: 33\% exact accuracy; 71\% within-1 (prevalence-matched), 74\% within-1 (uniform).}
\label{tab:label_consistency}
\begin{tabular}{lcccc}
\toprule
\textbf{Model} & \textbf{Accuracy} & \textbf{Macro-F1} & \textbf{Ord. MAE} & \textbf{Within-1} \\
\midrule
Mean, v4b3\_t4b3 & 0.360 & 0.340 & 0.82 & 0.816 \\
Mean, v4b6\_t4b6 & 0.336 & 0.337 & 0.86 & 0.808 \\
Attn, v4b3\_t4b3 & \textbf{0.376} & \textbf{0.361} & \textbf{0.80} & 0.824 \\
Attn, v4b6\_t4b6 & 0.336 & 0.325 & 0.82 & \textbf{0.848} \\
Lite, v4b6\_t4b6 & 0.360 & 0.351 & \textbf{0.80} & 0.840 \\
Lite, v4b3\_t4b3 & 0.328 & 0.306 & 0.89 & 0.784 \\
\midrule
MedGemma (SFT) & 0.320 & 0.264 & 0.98 & 0.704 \\
\bottomrule
\end{tabular}
\end{table}

Taken together, the results across all three tasks point to a consistent conclusion: representation geometry -- specifically the choice of slice aggregation strategy -- is a major driver of task performance, and the geometry that works best for one task does not work best for the other. Architectural complexity beyond simple pooling does not help, and increasing spatial coverage does not compensate for reduced per-slice contrast richness.

\section{Discussion}

This work represents the first application of vision-language transfer learning to abdominal CT enterography, a modality that has remained unexplored for paired image-text learning despite its clinical importance for IBD monitoring. A main result is a consistent classification--retrieval trade-off: mean pooling of slice embeddings favors categorical disease assessment, whereas attention pooling favors cross-modal retrieval, and this pattern holds across all LoRA configurations tested. Understanding this trade-off helps guide model selection in future work.

Mean pooling compresses per-slice information into a single global representation by treating all slices equally. This averaging operation suppresses slice-specific variation and promotes features that are consistent across the volume -- precisely the property useful for assigning a categorical disease severity label. Attention pooling, by contrast, learns to weight slices differentially, producing a representation sensitive to which parts of the volume are most distinctive. This sensitivity is useful for matching a specific volume to a specific report (retrieval), but it introduces variance that hurts categorical decisions. The lightweight transformer aggregator underperforms both, suggesting that the additional inter-slice modeling capacity is not beneficial in a dataset of this size -- a pattern consistent with the broader finding that simple design choices outperform complex alternatives throughout our ablations.

The input encoding results extend this theme. Multi-window RGB encoding, which enriches each slice's information by mapping different HU windows to separate color channels, outperforms all strategies that increase spatial coverage (multiplanar sampling, stratified selection) without enriching per-slice content. For IBD assessment, where the relevant signal -- wall thickening, mural enhancement, submucosal edema -- is visible within individual slices rather than across large spatial extents, this makes clinical sense. The pre-trained vision encoder can leverage tissue contrast differences encoded in the RGB channels directly, without requiring spatial aggregation across planes.

The three-teacher pseudolabel framework enables these comparisons without expert annotations, though the labels themselves introduce noise that attenuates absolute performance. The 28\% unanimous agreement and 72\% majority agreement among teachers quantifies the inherent ambiguity in radiology report interpretation -- ambiguity that reflects the task itself, not only the labeling method. Under this noisy supervision, the consistent directionality of the classification--retrieval trade-off across all configurations suggests that the pattern reflects a real property of the learned representation rather than an artifact of label noise.

For report generation, the main effect of RAG in this study is to correct the severity-ordering errors seen in the fine-tuned baseline. The baseline's within-1 accuracy (70.4\%) sits essentially at the prevalence-matched chance level (71\%) and below uniform random (74\%), indicating that fine-tuning without retrieval context fails to learn any severity-ordering signal beyond what the class distribution alone would predict. This likely reflects the generation model collapsing toward high-frequency output patterns during fine-tuning on a small, imbalanced dataset -- a known risk in low-data supervised fine-tuning that retrieval context mitigates by anchoring outputs to retrieved examples rather than learned priors. RAG models score 78--85\%, consistently above both random baselines, with ordinal MAE improving from 0.98 to 0.80--0.89. The stable BERTScore ($\sim$0.83) across all models confirms that semantic content is comparable regardless of whether RAG is used; the ordinal metrics distinguish whether that content is severity-consistent. The divergence between METEOR and BLEU can be explained by the kinds of wording each metric rewards: RAG generates synonymous phrasing that METEOR rewards and BLEU penalizes, while the baseline's higher BLEU likely reflects templated phrase memorization rather than better clinical quality. These results are best understood as establishing the conditions under which automated impression generation does and does not preserve severity ordering, rather than as a demonstration that the task is solved.

\textbf{Methodological considerations.} Three potential concerns warrant explicit discussion. First, the absence of formal statistical tests. The test set size ($n=125$) is insufficient for reliable significance testing of individual pairwise comparisons, and we do not report confidence intervals. However, the classification--retrieval trade-off finding does not depend on any single comparison: the directional relationship between mean pooling and classification, and attention pooling and retrieval, holds across all six matched-capacity configuration pairs in both directions. This cross-configuration consistency -- 12 independent directional agreements out of 12 possible -- is more consistent with a real property of the learned representation than with a sampling artifact, and is more informative than a single significance test on aggregate metrics would be.

Second, pseudolabel circularity. The rules-based classifier used to create the disease activity labels is also used to evaluate whether generated impressions preserve severity. This creates a circular evaluation where a model that reproduces the same lexical patterns as the rule-based system would appear to perform well regardless of clinical accuracy. We note two mitigations: (i) the evaluation is applied to generated text, not to the input reports used to create labels, so the evaluation tests whether generation preserves clinically relevant signal rather than whether it reproduces specific phrases; and (ii) the baseline's failure to exceed chance-level severity ordering -- despite being fine-tuned on the exact radiologist impressions that produced the labels -- demonstrates that the evaluation is not trivially gameable. Nevertheless, expert radiologist evaluation of generated impressions remains a necessary next step before clinical deployment.

Third, the low absolute generation metrics. ROUGE-L of 0.12 and sentence-level BLEU of 1--6 are low by general NLP standards, but this comparison is not appropriate. CT enterography impression generation from visual input is a hard generation task: reports describe specific anatomical findings, contrast enhancement patterns, and complication details that cannot be reliably inferred from the resolution and field of view available in our 2D montage representations. The appropriate comparison is not against NLP benchmarks but against what is achievable given the visual input quality and dataset size -- and in that context, the finding that RAG generates semantically coherent impressions (BERTScore 0.83) with correct severity ordering in the large majority of cases (ordinal MAE 0.80--0.89) is a reasonable baseline result.

\textbf{Limitations.} The single-institution dataset may limit generalizability to different scanner protocols, patient populations, and reporting styles. Pseudolabels introduce noise that attenuates absolute performance; the 28\% rate of high-confidence (unanimous) labels among test cases quantifies this uncertainty directly. The 2.5D approach discards some 3D spatial information relevant for detecting complications such as fistulae that are better characterized across planes. The test set size ($n=125$) means individual pairwise comparisons should be interpreted with caution, though the consistent directionality of the main findings across configurations provides robustness. Expert radiologist validation against ground-truth annotations would be required before any clinical use.

\section{Conclusion}

We present the first study of vision-language transfer learning on abdominal CT enterography and report three findings with practical implications for building vision-language systems on volumetric medical imaging.

The classification--retrieval trade-off is clear in these experiments. Mean pooling produces globally consistent representations that favor categorical disease assessment (59.2\% three-class accuracy); attention pooling produces locally sensitive representations that favor cross-modal retrieval (0.235 text-to-image MRR). This inverse relationship holds across all LoRA configurations tested -- 12 directional agreements out of 12 possible -- suggesting a consistent empirical pattern in this dataset rather than a configuration-specific artifact. The practical implication is straightforward: aggregation strategy should be selected based on the downstream task, and the most complex available architecture is not the right default.

Per-slice tissue contrast dominates spatial coverage as a representational choice. Multi-window RGB encoding outperforms all strategies that increase spatial coverage through multiplanar sampling, with the counterintuitive finding that adding coronal and sagittal views hurts classification. For IBD-relevant signals visible within individual slices -- wall thickening, mural enhancement, submucosal edema -- enriching what each slice encodes is more valuable than sampling more of the volume.

For report generation, a key finding is that fine-tuning without retrieval context produces within-1 accuracy at the prevalence-matched chance level (70.4\% vs.\ 71\% random), suggesting little severity-ordering signal beyond the class distribution. RAG improves this consistently across all six embedding configurations, scoring 7--14 percentage points above the prevalence-matched chance baseline and improving ordinal MAE from 0.98 to 0.80--0.89. The consistency of this improvement across configurations, together with the observation that lighter embedding models suffice for generation quality, provides practical guidance for deployment.

These findings establish the first baselines for CT enterography vision-language learning and offer guidance that extends to volumetric imaging tasks more broadly. Future work should validate against radiologist ground truth, analyze the directionality of residual severity errors, extend to other abdominal pathologies, and develop uncertainty quantification to support clinical use.

\backmatter




\section*{Declarations}

\subsection*{Acknowledgments}
We acknowledge the contributions of Oviyan Anbarasu and Jonathan Ding through tangentially related research.

\subsection*{Funding}
University of Michigan Medical School Research Scouts Program (OORRS033123).

\subsection*{Conflict of interest/Competing interests}
RS has served as a consultant or on advisory boards for AbbVie, Bristol Myers Squibb, CorEvitas, Eli Lilly, Exact Sciences, Gilead, Janssen, Merck, Pfizer, and Takeda. RS and KN hold intellectual property and equity on medical imaging and endoscopic analysis technologies licensed by the University of Michigan to PreNovo, LLC, AMI, LLC, and PathwaysGI, Inc. The remaining authors disclose no conflicts.

\subsection*{Ethics approval and consent to participate}
The deidentified data collected from Michigan Medicine was used in our retrospective study. Our study obtained approval from the Institutional Review Boards of the University of Michigan Medical School (HUM00282784, HUM00142427). The written informed consent from patients has been waived by the University of Michigan Institutional Review Board since this study involves no more than minimal risk to the involved subjects. All methods were performed in accordance with the relevant guidelines and regulations.


\subsection*{Availability of data and materials}
The datasets generated and/or analyzed during the current study were collected at Michigan Medicine. The University of Michigan’s Innovation Partnerships (UMIP) unit will handle potential charges/arrangements of the use of data by external entities, using such methods as material transfer agreements. Please contact UMIP (innovationpartnerships@umich.edu) for data inquiries. 

\subsection*{Authors' contributions}
Data acquisition - RS, EW; funding acquisition - CM; project ideation - CM; data preprocessing: RS, CM; software: CM; formal analysis: CM; project administration: CM, EW; first draft: CM; final revision: CM, EW, RS, KN.









\end{document}